\documentclass[11pt,a4paper]{article}
\usepackage[english]{babel}
\usepackage[hyperref]{acl2017}
\usepackage{times}
\usepackage{latexsym}

\usepackage[hang]{footmisc}
\usepackage{hyperref}
\usepackage{amsmath}
\usepackage{amssymb}
\usepackage{bbm}
\usepackage{multirow}
\usepackage{graphicx}
\usepackage[linesnumbered,ruled]{algorithm2e}
\usepackage{footmisc}
\usepackage{float}

\usepackage{rotating}
\newcommand{\COPY}{\begin{turn}{90}{COPY~}\end{turn}}
\newcommand{\DEL}{\begin{turn}{90}{DELETE$\;$}\end{turn}}
\newcommand{\STEP}{\begin{turn}{90}STEP$\;$\end{turn}}
\newcommand{\STOP}{\begin{turn}{90}STOP\end{turn}}

\aclfinalcopy 

\DeclareMathOperator*{\argmax}{arg\,max}
\DeclareMathOperator*{\nbest}{n-best}

\newcommand{\bv}[1]{\mathbf{#1}}

\usepackage{xspace}
\newcommand{\HACM}{HACM\xspace}
\newcommand{\HAEM}{HAEM\xspace}
\newcommand{\CM}{\textrm{CM\xspace}}
\newcommand{\EM}{\textrm{EM\xspace}}

\title{Align and Copy: UZH at SIGMORPHON 2017 Shared Task for Morphological Reinflection}
\author{Peter Makarov$^{\dagger}$\thanks{These two authors contributed equally.} \hspace{5mm} Tatiana Ruzsics$^{\ddagger}$\footnotemark[1] \hspace{5mm} Simon Clematide$^{\dagger}$ \\ 
$^{\dagger}$Institute of Computational Linguistics, University of Zurich, Switzerland \\
$^{\ddagger}$CorpusLab, URPP Language and Space, University of Zurich, Switzerland \\
{\small {\tt makarov@cl.uzh.ch \hspace{2mm} tatiana.ruzsics@uzh.ch \hspace{2mm} simon.clematide@cl.uzh.ch}} }

\date{}

\begin{document}
\maketitle
\begin{abstract}
This paper presents the submissions by the University of Zurich to 
the SIGMORPHON 2017 shared task on morphological reinflection. The task is to predict the inflected form given a lemma and a set of morpho-syntactic features. We focus on neural network approaches that can tackle the
task in a limited-resource setting. As the transduction of the lemma into the
inflected form is dominated by copying over lemma characters, we propose two
recurrent neural network architectures with hard monotonic attention that are
strong at copying and, yet, substantially different in how they achieve this. The first approach is an encoder-decoder model with a copy mechanism.
The second approach
is a neural state-transition system over a set of explicit edit actions,
including a designated COPY action. We experiment with character alignment and find that naive, greedy alignment consistently produces strong results for some languages. Our best system combination is the overall winner of the SIGMORPHON 2017 Shared Task 1 without external resources.  At a setting with 100 training samples, both our approaches, as ensembles of models, outperform the next best competitor.

\end{abstract}

\section{Introduction}
This paper describes our approaches and results for Task~1 (without external resources) of the CoNLL-SIGMORPHON 2017 challenge on Universal Morphological Reinflection \cite{cotterelletal2017}. This task consists in generating inflected word forms for 52 languages given a lemma and a morphological feature specification \cite{Sylak-GlassmanKirov:2015} as input (Figure~\ref{fig:morphexample}). 

\begin{figure}[h]
\centering
\resizebox{0.8\linewidth}{!}{
\begin{tabular}{ccc}
fliegen & \multirow{2}{*}{\scalebox{2}{$\Huge\rhd$}} & \multirow{2}{*}{flog} \\
$\{$\textsc{Verb, Past Tense,} & & \\
\textsc{3rd Person, Singular}$\}$ & & 
\end{tabular}}
\caption{Morphological inflection generation task. A German language example.}
\label{fig:morphexample}
\end{figure}
There are three task setups: a low setting where training data are only 100 (!) samples, a medium setting with 1K training samples, and a high setting with 10K samples.
We consider the problem of tackling morphological inflection generation at a low-resource setting with a neural network approach, which is hard for plain soft-attention encoder-decoder models \cite{Kann&Schutze2016a,Kann&Schutze2016b}. 
We present two systems that are based on the hard monotonic attention model of \newcite{Aharoni&Goldberg2017,Aharonietal2016}, which is strong on smaller-sized training datasets.
We observe that to excel at a low-resource setting, a model needs to be good at copying lemma characters over to the inflected form---by far the most common operation of string transduction in the morphological inflection generation task.

In our first approach, we extend the hard monotonic attention model with a copy mechanism that produces a mixture distribution from the character generation and character copying distributions. This idea is reminiscent of the pointer-generator model of \newcite{Seeetal2017} and the CopyNet model of \newcite{GuCopyNet:2016}.

Our second approach is a neural state-transition system that explicitly learns the copy action and thus does away with character decoding altogether whenever a character needs to be copied over. This approach is inspired by shift-reduce parsing with stack LSTMs \cite{Dyeretal2015} and transition-based named entity recognition \cite{Lampleetal2016}.

\section{Preliminaries}
In this section, we formally describe the problem of morphological inflection generation as a string transduction task. Next, we show how this task can be reformulated in terms of transduction actions. Finally, we discuss the string alignment strategies that we use to derive oracle actions.

\subsection{Morphological inflection generation}
Morphological inflection generation is an instance of the more general sequence transduction task, where the goal is to find  
a mapping of a variable-length sequence $x$ to another variable-length sequence $y$. Specific to morphological inflection generation is that the input and output vocabularies---lemmas and inflected forms---are the same set of characters of one natural language, i.e. $\Sigma^{x} = \Sigma^{y} = \Sigma$. Formally, our task is to learn a mapping from an input sequence of characters $x_{1:n} \in \Sigma^{*}$ (the lemma) to an output sequence of characters $y_{1:m} \in \Sigma^{*}$ (the inflected form) given a set of morpho-syntactic features $f \subseteq \Phi$, where $\Phi$ is the alphabet of morpho-syntactic features for that language. 

\subsection{Task reformulation}
To efficiently condition on parts of the input sequence, we use hard monotonic attention, which has been found highly suitable for this task \cite{Aharoni&Goldberg2017,Aharonietal2016}. With \emph{hard attention}, at each step, the prediction of an output element is based on attending to only one element from the input sequence as opposed to conditioning on the entire input sequence as in soft attention models. 

\emph{Hard monotonic attention} is motivated by the often monotonic alignment between the lemma characters and the characters of its inflected form: It suffices to only allow for the advancement of the attention pointer up in a sequential order over the elements of the input sequence.
Thus, the sequence transduction process can be represented as a sequence of actions $a_{1:q} \in \Omega^*$ over an input string, where the set of actions $\Omega$ includes operations for writing characters and advancing the attention pointer. We can, therefore, reformulate the task of finding a mapping from an input sequence of lemma characters
$x \in \Sigma^*$ to the output sequence of actions $\hat{a} \in \Omega^*$, given a set of morpho-syntactic features $f \subseteq \Phi$, such that:
\begin{align}
\hat{a} =& \argmax_{a \in \Omega^*} P\big( a 
\mid x, f \big) \nonumber \\
\label{eq:obj}
	=& \argmax_{a \in \Omega^*} \prod_{t = 1}^{|a|} P \big( a_t \mid a_{1:t-1}, x, f \big)
\end{align}
We use a recurrent neural network to estimate the probability distribution $P$ in Equation~\ref{eq:obj} from training data. To derive the sequence of oracle actions from each training sample, we use two different character alignment strategies formally described below.

\subsection{Character alignment strategies}
\label{sec:align}
We use two string alignment strategies that produce 0-to-1, 1-to-0, and 1-to-1 character alignments (Figure~\ref{fig:align}).
\paragraph{Smart} alignment uses the Chinese Restaurant Process character alignment  implementation distributed with the SIGMORPHON 2016 baseline system \cite{CotterellKirov:2016}.\footnote{\url{https://github.com/ryancotterell/sigmorphon2016/tree/master/src/baseline}} 
This is the aligner of \newcite{Aharoni&Goldberg2017}. 
\paragraph{Naive} 
alignment aligns two sequences $p$ and $q$, such that the length of $p$ is greater or equal to the length of $q$, by producing 1-to-1 character alignments until it reaches the end of $q$, from which point it outputs 1-to-0 alignments (and 0-to-1 alignments once it reaches the end of $p$ if $|q| > |p|$). 
%

\begin{figure}[t]
\centering
\begin{tabular}{llllllll}
f & l & o &   & g &  &  \\
$\vert$ & $\vert$ & $\vert$ & $\vert$ & $\vert$ & $\vert$ & $\vert$ \\
f & l & i & e & g & e & n  \\
\end{tabular}

\bigskip
\begin{tabular}{llllllll}
f & l & o  & g &  &  &  \\
$\vert$ & $\vert$ & $\vert$ & $\vert$ & $\vert$ & $\vert$ & $\vert$ \\
f & l & i & e & g & e & n  \\
\end{tabular}
\caption{Examples of smart alignment (top) and naive alignment (bottom). In each example, inflected form is at the top, lemma at the bottom.}
\label{fig:align}
\end{figure}

\begin{figure*}[t]
\begin{center}
\includegraphics[width=0.9\textwidth]{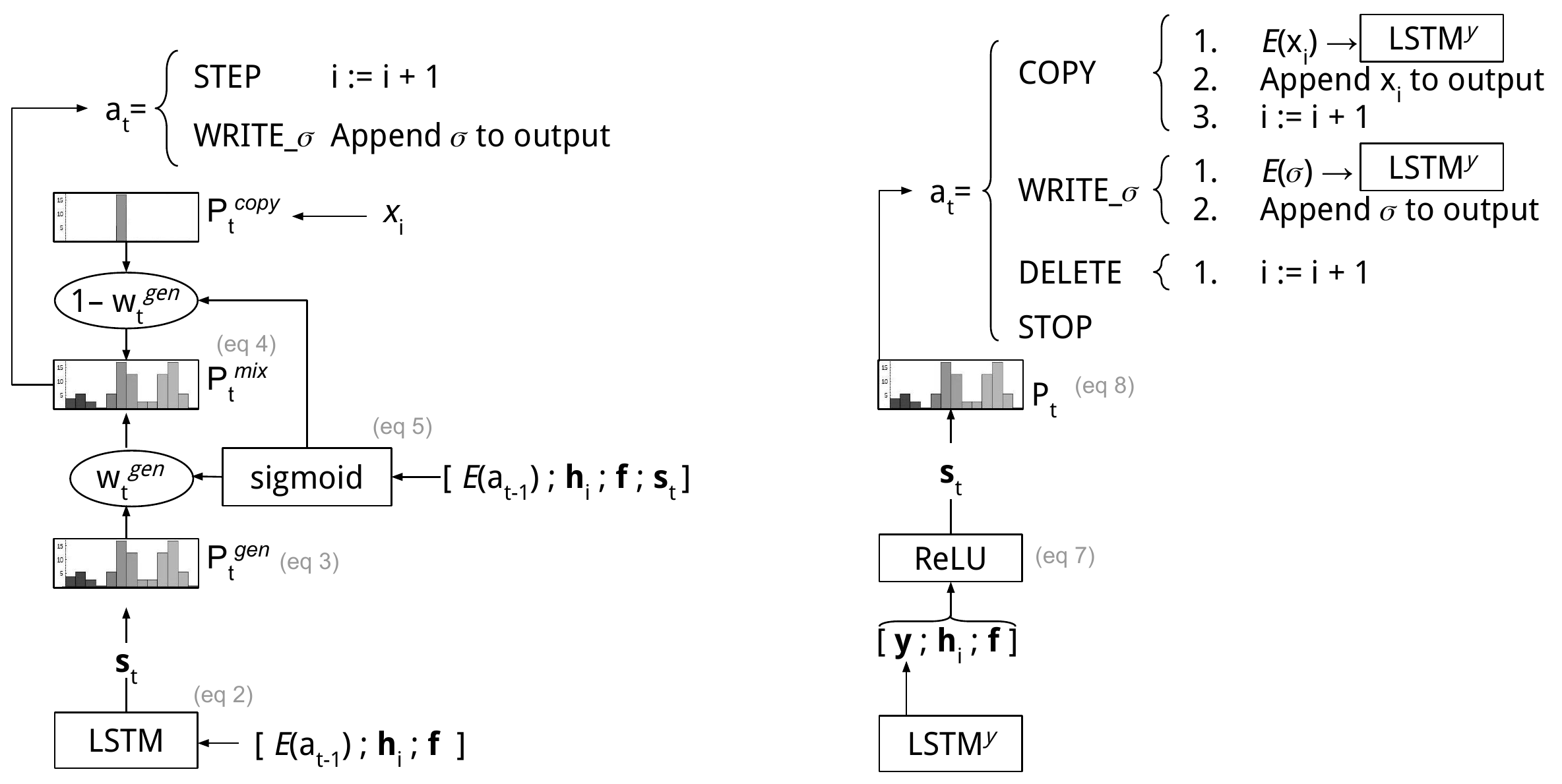}
\caption{Overview of the architectures. Hard attention model with copy mechanism (\HACM) on the left, hard attention model over edit actions (\HAEM) on the right.}
\label{fig:sigmorphon-systems-diagram}
\end{center}
\end{figure*}

\section{First approach: Hard attention model with copy mechanism (\HACM)}

Our first approach augments the hard monotonic attention model of \newcite{Aharoni&Goldberg2017} with a copy mechanism which adds a soft switch between generating an output symbol from a fixed vocabulary $\Sigma^{train}$ and copying the currently attended input symbol $x_i$. In this section, we first review the architecture of the hard monotonic model and then present our copy mechanism.

\subsection{Hard monotonic attention model}

The hard monotonic attention model operates over two types of actions: WRITE\_$\sigma$, $\sigma \in \Sigma$, for outputting the character $\sigma$ and STEP which moves forward the attention pointer, i.e. $\Omega = \Sigma \cup \{ \textrm{STEP} \}$. At each step, the model either generates an output symbol
or starts to attend to the next encoded input character. 
The system learns to move the attention pointer by outputting a STEP action.
To compute the sequences of oracle actions for each training pair of lemma and its inflected form, \newcite{Aharoni&Goldberg2017} apply a deterministic algorithm\footnote{We refer the reader to \newcite{Aharoni&Goldberg2017} for the description of the algorithm.} to the output of the smart aligner.

\paragraph{Architecture}

The hard monotonic attention model
uses a single-layer bidirectional LSTM encoder \cite{Graves&Schmidhuber2005} to encode input lemma $x_{1:n}$ as a sequence of vectors $\bv{h}_{1:n}, \bv{h}_i \in \mathbb{R}^{2H}$, where $H$ is the hidden dimension of the LSTM layer.

At all time steps $t$, the model maintains a state $\bv{s}_t \in \mathbb{R}^{H}$ from which the most probable action $a_t$ is predicted. The sequence of states is modeled with a single-layer LSTM that receives, at time $t$, a concatenated input of:
\begin{enumerate}
\item the currently attended vector $\bv{h}_i \in \mathbb{R}^{2H}$, where $i$ is the attention pointer,

\item the concatenated vector of feature embeddings $\bv{f} \in \mathbb{R}^{F \cdot \vert \Phi \vert}$, where $F$ is the dimension of the feature embedding layer,

\item the embedding of the previous output action $E(a_{t-1}) \in \mathbb{R}^E$, where $E$ is the dimension of the action embedding layer.
\end{enumerate}

\begin{equation}
\label{eq:lstm}
\bv{s}_t = \textrm{LSTM}\big( [E(a_{t-1}) ; \bv{h}_i ; \bv{f} ] \big)
\end{equation}
Let $\Sigma^{train} \subseteq \Sigma$ be the set of characters in training data. Then, the distribution $P_t^{gen} $ for generating actions over the vocabulary $\Omega^{train}=\Sigma^{train}\cup \{ \textrm{STEP} \}$ is modeled with the softmax function:

\begin{equation}
\label{eq:pgen}
P_t^{gen} = \textrm{softmax}\big( \bv{W} \cdot \bv{s}_t + \bv{b} \big)
\end{equation}

When the predicted action is STEP, the attention index gets incremented $i := i+1$, and so at the next time step $t+1$, the model attends to vector $h_{i+1}$ of the bidirectionally encoded lemma sequence.

\begin{table*}[tb]
\centering
\resizebox{\textwidth}{!}{
\begin{tabular}{cccccccccccccc||l}
1 & 2 & 3 & 4 & 5 & 6 & 7 & 8 & 9 & 10 & 11 & 12 & 13 & 14 & $t$ \\
$\langle$s$\rangle$ & & f & & l &  & o & & & g &  &  &  & $\langle$/s$\rangle$ &  $y$ \\ \hline
$\langle$s$\rangle$ & \STEP & f & \STEP & l & \STEP & o & \STEP & \STEP & g & \STEP & \STEP & \STEP & $\langle$/s$\rangle$ & $a_{t}$ \\ \hline
$\langle$s$\rangle$ & $\langle$s$\rangle$ & f & f & l & l & i & i & e & g & g & e & n & $\langle$/s$\rangle$ & $x_i$ \\
0 & 0 & 1 & 1 & 2 & 2 & 3 & 3 & 4 & 5 & 5 & 6 & 7 & 8 &  $i$
\end{tabular}
\qquad
\begin{tabular}{ccccccccc||l}
1 & 2 & 3 & 4 & 5 & 6 & 7 & 8 & 9 & $t$ \\
f & l &   &   & o & g &   &   &   & $y$ \\ \hline
\COPY & \COPY & \DEL & \DEL & o & \COPY & \DEL & \DEL & \STOP & $a_{t}$ \\ \hline
f & l & i & e & g & g & e & n & -- & $x_{i}$ \\
1 & 2 & 3 & 4 & 5 & 5 & 6 & 7 & 7 & $i$
\end{tabular}}
\caption{Examples of generating German ``flog'' from ``fliegen'': \HACM (left), \HAEM (right). $i$ is the attention pointer, $x_i$ the currently attended lemma character, $a$ the sequence of actions, $y$ the output, $t$ the index over actions.}
\label{tab:acts}
\end{table*}

\subsection{Copy mechanism}
Our copying mechanism is based on using a mixture of a \emph{generation} probability distribution from Equation~\ref{eq:pgen} and a \emph{copying} probability distribution.
At each time step $t$, the action $a \in \Omega^{train}$ is predicted from the following mixture distribution:
\begin{equation}
\label{eq:mix}
P_t (a) = w_t^{\textrm{gen}}P_t^{gen} (a) + (1-w_t^{\textrm{gen}}) \mathbbm{1}_{\{ a=x_i\}},
\end{equation}
where $x_i$ is the currently attended character of the lemma sequence $x$ and $\mathbbm{1}_{\{ a=x_i\}} = P^{copy}_t (a)$ is a probability distribution for copying $x_i$.

The mix-in parameter of the generation distribution $w_t^{\textrm{gen}} \in \mathbb{R}$ is calculated from the concatenation of the state vector $\bv{s}_t$ and the input vector that produces this state. The resulting vector is fed through a linear layer to the logistic sigmoid function:
\begin{equation}
w_t^{\textrm{gen}}= \textrm{sigmoid}(\bv{v} \cdot [\bv{h}_i;\bv{f};E(a_{t-1});\bv{s}_t] + c)
\end{equation}
The mix-in parameter serves as a switch between a) generating a character from $\Sigma^{train}$ according to the generation distribution $P_t^{gen}$, and b) copying the currently attended character $x_i \in \Sigma^{train}$.

At test time, we allow the copying of out-of-vocabulary (OOV) symbols by adding the following modification to the mixture distribution in Equation~\ref{eq:mix}:
\begin{equation}
\begin{aligned}
\label{eq:mix_test}
P_t (a) =& \mathbbm{1}_{\{ a=x_i\}}\mathbbm{1}_{\{ x_i \in \Sigma \setminus \Sigma^{train} \}} + \bigg( w_t^{\textrm{gen}}P_t^{gen} (a) \\ &+ (1-w_t^{\textrm{gen}})\mathbbm{1}_{\{ a=x_i\}} \bigg) \mathbbm{1}_{\{ x_i\in \Sigma^{train}\}} 
\end{aligned}
\end{equation}
Therefore, if the currently attended symbol $x_i$ is OOV, we copy it with probability one according to the distribution $\mathbbm{1}_{\{ a=x_i\}}$; otherwise, we use the mixture of generation $P_t^{gen}$ and copy $\mathbbm{1}_{\{ a=x_i\}}$ distributions. Thus, the distribution $P_t$ is built over an instance specific vocabulary 
${\Omega^{train}}\cup \{x_i\}$.
After copying the OOV symbol, we advance the attention pointer and use STEP as the previous predicted action.

The full architecture of the \HACM system is shown schematically in Figure~\ref{fig:sigmorphon-systems-diagram}.

\subsection{Learning}
We train the system using cross-entropy loss, which, for a single input $(x, y, f)$, equates to:
\begin{equation}
\label{eq:loss}
\mathcal{L}(\Theta ; x, a, f) = - \sum_{t = 1}^{|a|} \log P_t \big(a_t \mid a_{1:t-1}, x, f \big),
\end{equation}
where $x, y$ are lemma and inflected form character sequences, $f$ the set of morpho-syntactic features, $a$ the sequence of oracle actions derived from $(x, y)$, $\Theta$ the model parameters and $P_t$ is the probability distribution over actions from Equation~\ref{eq:mix}.

\section{Second approach: Hard attention model over edit actions (\HAEM)}

This neural state-transition system also uses hard monotonic attention but transduces the lemma into the inflected form by a sequence of explicit \emph{edit} actions: COPY, DELETE, and WRITE\_$\sigma$, $\sigma \in \Sigma$. The architectures of the two models are also different (Figure~\ref{fig:sigmorphon-systems-diagram}).

\subsection{Semantics of edit actions}

\paragraph{COPY} If the system generates COPY, the lemma character at the attention index $x_i$ is appended to the current output of the inflected form and the attention index is incremented $i := i + 1$. Therefore, unlike other neural morphological inflection generation systems, the copy character is not decoded from the neural network.

\paragraph{DELETE} The system generates DELETE if it needs to increment the attention index.

\paragraph{WRITE\_$\sigma$} Whenever the system chooses to append a character $\sigma \in \Sigma$ to the current output of the inflected form, such that $\sigma \neq x_i$ where $x_i$ is the lemma character at the attention index, it generates the corresponding WRITE\_$\sigma$ action.

Using this set of edit actions, the system can copy, delete, and substitute new characters. The substitution of a new character $\sigma$ for a currently attended lemma character $x_i$, $\sigma \neq x_i$, is expressed as a sequence of one DELETE and one WRITE\_$\sigma$ action.

This action set directly compares to the $\Omega = \Sigma \cup \{\rm{STEP}\}$ actions of the \HACM model,
which uses most basic actions to express edit operations. Crucially, in the \HAEM system,
character copying is a single action (which does not require character decoding) whereas it is typically a sequence of one WRITE\_$\sigma$ (=$\sigma$) and one STEP action in \HACM.%
\footnote{Except whenever the next alignment is 0-to-1 the \HACM does not generate STEP. The \HAEM system, however, increments the attention index on every COPY action.}
Further, \HAEM effectively deals with OOV characters through COPY and DELETE actions.

\paragraph{STOP} Additionally, to signal the end of transduction, the system generates a STOP action.

\subsection{Deriving oracle actions}
We use the character alignment methods of Section~\ref{sec:align} to deterministically compute sequences of oracle actions for each training example using Algorithm~\ref{alg:1}.
\begin{algorithm}
    \SetKwInOut{Input}{Input}
    \SetKwInOut{Output}{Output}
   \Input{$A$, list of 1-to-1, 0-to-1, and 1-to-0 alignments between lemma and form}
   \Output{$O$, list of oracle actions }
   \ForEach{$(t, s) \in A$}{
    \uIf{$t = \epsilon$}{
     $O$.append(WRITE\_$s$)
     }
     \uElseIf{$s = \epsilon$}{
           $O$.append(DELETE)

     }
     \uElseIf{$s = t$}{
          $O$.append(COPY)

     }
     \Else{
        $O$.append(DELETE)
        
        $O$.append(WRITE\_$s$)
    }
     
    }       
    \caption{Derivation of oracle actions from alignment of lemma and  form.}
    \label{alg:1}
\end{algorithm}

We then normalize all sub-sequences of only DELETE and WRITE\_$\sigma$ in such a way that all DELETEs come before all WRITE\_$\sigma$ actions. This simplifies unintuitive alignments produced by the smart aligner, especially at the low setting.

\subsection{Architecture}
Similarly to \HACM, the input lemma is encoded as a sequence of vectors $\bv{h}_{1:n}, \bv{h}_i \in \mathbb{R}^{2H}$ with a single-layer bidirectional LSTM. Additionally, we use a single-layer LSTM to represent the predicted inflected form $y_{1:m}$, to which we refer as $\rm{LSTM}^y$. In case the model outputs a character with WRITE\_$\sigma$ or COPY, $\rm{LSTM}^y$ gets updated with the embedding of this character.

At all time steps $t$, the system maintains a state $\bv{s}_t \in \mathbb{R}^{H}$ from which it predicts the most probable action $a_t$. The state sequence is derived differently. At time $t$, a concatenation of:
\begin{enumerate}
\item the currently attended vector $\bv{h}_i \in \mathbb{R}^{2H}$,

\item the set-of-features vector $\bv{f} \in \mathbb{R}^{ \vert \Phi \vert}$,

\item the output of the latest state $\bv{y} \in \mathbb{R}^{H}$ of the inflected form representation $\rm{LSTM}^y$,
\end{enumerate}
passes through a rectifier linear unit (ReLU) layer \cite{Glorotetal2011} to finally produce the state vector $\bv{s}_t$.

The probability distribution over valid actions%
\footnote{Some actions are not valid in certain states: The system cannot DELETE or COPY if the attention index is greater than the length of the lemma.} is then computed with softmax:
\begin{align}
\bv{s}_t &= \rm{ReLU}\big( \bv{W} \cdot [\bv{y} ; \bv{h}_i ; \bv{f}] + \bv{b}  \big)
\\
P_t &= \textrm{softmax}\big(\bv{V} \cdot \bv{s}_t + \bv{c} \big)
\end{align}

This describes the basic form of the \HAEM system (Figure~\ref{fig:sigmorphon-systems-diagram}). In our experiments, we extend it to include two more representations: an LSTM that represents the action history, $\rm{LSTM}^a$, and another LSTM that encodes a sequence of deleted lemma characters, $\rm{LSTM}^d$. The deletion $\rm{LSTM}^d$ gets emptied once a WRITE\_$\sigma$ action is generated. In this way, we attempt to keep in memory a full representation of some sub-sequence of the lemma that needs to be replaced in the inflected form. In the extended system, the state $\bv{s}_t$ is thus derived from an input vector $[\bv{y} ; \bv{h}_i ; \bv{f} ; \bv{a} ; \bv{d}]$, where $\bv{a} \in \mathbb{R}^{H}$ is the output of the latest state of the action history $\rm{LSTM}^a$ and $\bv{d} \in \mathbb{R}^{H}$ the output of the latest state of the deletion $\rm{LSTM}^d$.

The system is trained using the cross-entropy loss function as in Equation~\ref{eq:loss}.

\section{Experimental setup}

\begin{table}[tb]
\centering
\begin{tabular}{l|cc|cc|c}
system    &	\multicolumn{2}{c|}{\HACM} & \multicolumn{2}{c|}{\HAEM}&Nematus  \\
alignment &	S 	& 	N 	  &	S 	& 	N   &--  \\ \hline\hline
low       &    5&5	  &  5&5   \\
medium    &	5&5	  &	5	&	3 &     \\
high      &	3&3	 &	3	&	2   &1  \\ \hline
\end{tabular}
\caption{Number of single models that we train for each language. N=Naive alignment, S=Smart alignment. E.g. for each language at the medium setting, there are 3 \HAEM models trained on data aligned with naive alignment.}
\label{tab:num-models}
\end{table}

\begin{table}[tb]
\centering
\resizebox{\linewidth}{!}{
\begin{tabular}{l|ll}
Run & Systems & Strategy \\ \hline \hline
1 & \multirow{2}{*}{\CM} & MAX $\{$ E(N$_{\CM}$), E(S$_{\CM}$) $\}$ \\
2 & & ENSEMBLE\_$7$ (N$_{\CM}$ $\cup$ S$_{\CM}$) \\ \hline
3 & \multirow{2}{*}{\EM} & MAX $\{$ E(N$_{\EM}$), E(S$_{\EM}$) $\}$ \\
4 & & ENSEMBLE\_$7$ (N$_{\EM}$ $\cup$ S$_{\EM}$) \\ \hline
5 & \multirow{3}{*}{\CM \& \EM} & MAX $\{$ E(N$_{\CM}$), E(S$_{\CM}$), E(N$_{\EM}$), E(S$_{\EM}$)  $\}$ \\
6 &                         & ENSEMBLE\_$15$ (N$_{\CM}$ $\cup$ S$_{\CM}$ $\cup$ N$_{\EM}$ $\cup$ S$_{\EM}$) \\
7 &                         & MAX $\{$ Run 5, Run 6 $\}$ \\ \hline
\end{tabular}}

\caption{Aggregation strategies in submissions. \CM=\HACM, \EM=\HAEM, N$_{\CM}$=the set of \HACM models trained on naive-aligned data, S$_{\CM}$=the set of \HACM models trained on smart-aligned data, and similarly for \HAEM.}
\label{tab:runstrat}
\end{table}

We submit seven runs: a) two runs (1 and 2) for the \HACM model; b) two runs (3 and 4) for the \HAEM model; and c) three runs (5, 6, and 7) that combine both systems.
Detailed information on training regimes and the choice of hyperparameter values (e.g. layer dimensions, the application of dropout, etc.) for all the runs is provided in the Appendix. 
Crucially, for both systems and all settings and languages, we train models with both smart and naive alignments of Section~\ref{sec:align}. Table~\ref{tab:num-models} shows the number of single models for each system, setting, and alignment.\footnote{Due to time restrictions, we could not produce the target of 5 HAEM models per setting and alignment.} We decode using greedy search. 

We apply a simple post-processing filter that replaces any inflected form containing an endlessly repeating character with the lemma. This affects a small number of test samples---57 for \HACM and 238 for \HAEM across all languages and alignment regimes---and primarily at the low setting.

\begin{table*}[t]
\centering
\resizebox{\textwidth}{!}{
\begin{tabular}{lrr|rr||rr|rr|rrrr||rr}
 System &\multicolumn{2}{c|}{\HACM}  & \multicolumn{2}{c||}{\HAEM} & \multicolumn{2}{c|}{\HACM} & \multicolumn{2}{c|}{\HAEM} & \multicolumn{4}{c||}{\HACM \& \HAEM}  & BS & BS \\
Alignment/Run & N & S & N & S & 1 & 2 & 3 & 4 & 5 & 6 & 7 & 7 &  \\
Metric & Acc & Acc & Acc & Acc & Acc & Acc & Acc & Acc & Acc & Acc & Acc & Lev & Acc & Lev \\\hline
\multicolumn{15}{c}{Development Set}\\\hline
Low & 43.8 & 41.3 & 45.8 & 44.3 & 46.5 & 47.6 & 48.9 & 49.5 & 49.2 & 51.1 & \textbf{51.6} & 1.3 & \textit{38.0} & 2.1 \\

Medium & 75.8 & 81.4 & 70.7 & 80.0 & 81.9 & 82.6 & 80.5 & 81.1 & 82.2 & 83.4 & \textbf{83.5} & 0.3 & \textit{64.7} & 0.9 \\
High  & 93.3 & 94.6 & \textit{75.9} & 89.6 & 95.0 & 95.3 & 89.8 & 90.1 & 95.2 & 95.3 & \textbf{95.6} & 0.1 & 77.9 & 0.5 \\\hline
\multicolumn{15}{c}{Test Set}\\\hline
Low     &  &   &   &   & 46.0 & 46.8 & 48.0 & 48.5 & 48.2& \textbf{50.6} & \textbf{50.6} &1.3 & \textit{37.9} &  2.2\\

Medium &   &   &   &   &  80.9 & 81.8  &  79.6 & 80.3  & 81.0  &  \textbf{82.8}&   \textbf{82.8}& 0.3  &  \textit{64.7} & 0.9  \\
High   &   &  &    &   &  94.5 &  95.0 &  89.1 & 89.5 & 94.7  &  \textbf{95.1} &  \textbf{95.1} & 0.1 & \textit{77.8} & 0.5 \\\hline

\end{tabular}
}
\label{tab:results-average}
\caption{Macro average results over all languages for all settings on the official development and test set. N=Naive alignment, S=Smart alignment, BS=Baseline system, Acc=Accuracy, Lev=Levenshtein.}
\end{table*}
All runs aggregate the results of multiple single models, and we use a number of aggregation strategies. For system runs
1 through 4, these are:

\paragraph{Max strategy} For each language $l$, we compute two ensembles over single models---one ensemble $E(S)$ over smart alignment models and one ensemble $E(N)$ over naive alignment models. We then pick the ensemble with the highest development set accuracy for $l$:
\begin{equation}
\hat{M} = \argmax_{M \in \{ E(S), E(N) \}} \textrm{dev acc}(M)
\end{equation}
\paragraph{Ensemble\_{$n$} strategy} For each language $l$, we pick at most $n$ models from all single models such that they have the best development set accuracies for $l$. We then compute one ensemble over them:
\begin{equation}
\hat{M} =  E\bigg( 
\nbest_{M \in (S \cup N)}\;\textrm{dev acc}(M) \bigg)
\end{equation}

Runs 5, 6, and 7 are built with aggregation strategies that use as building blocks the MAX and ENSEMBLE\_$n$ strategies. Table~\ref{tab:runstrat} shows the strategies employed in each run.

At the high setting, Runs 5, 6, and 7 additionally feature a single run produced with Nematus \cite{Sennrichetal2017}, a soft-attention encoder-decoder system for machine translation. In all these runs, the Nematus run complements the HAEM models, which perform much worse at the high setting on average. We refer the reader to the Appendix for further information on data preprocessing, hyperparameter values, and training for the Nematus run.

\begin{figure*}[t]
\begin{center}
\includegraphics[width=\textwidth]{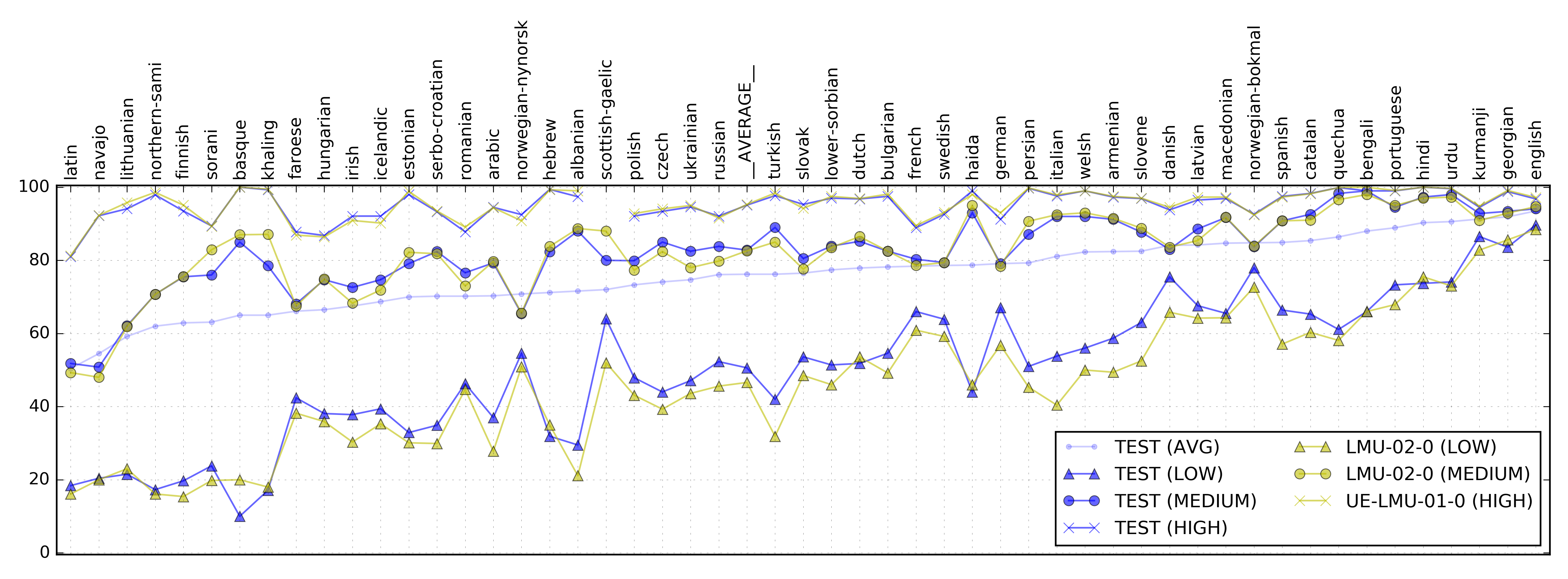}  
\caption{Test set accuracies of Run 7 (blue) and the next best system (yellow). The results are ordered by the averaged (low, medium, high) test set accuracies of Run 7.}
\label{fig:results-on-test-dev}
\end{center}
\end{figure*}

\section{Results and Discussion}
Table~\ref{tab:results-average} gives an overview on the average (macro) performance for each run on the official development and test sets at all settings. Accuracy measures the percentage of word forms that are inflected correctly (without a single character error). For the best system combination, we also report the average Levenshtein distance between the gold standard word form and the system prediction, which represents a softer criterion for correctness. Also, we include the performance of the shared task baseline system, which is a rule-based model that extracts prefix-changing and suffix-changing rules using alignments of each training sample with Levenshtein distance and associates the rules with the features of the sample.\footnote{\url{https://github.com/sigmorphon/conll2017/tree/master/baseline}} 
All our official runs beat the baseline by a large margin on average in terms of accuracy and also in terms of Levenshtein distance. For all settings, we see an improvement by applying the more complex ensembling strategies (Table~\ref{tab:runstrat}). It is the largest for low and the smallest for the high setting.



At the low setting, \HAEM outperforms \HACM on average by 2-3 percentage points accuracy and is, therefore, especially suited for a low resource situation. At the medium setting, the performance of \HACM is slightly better using smart alignments. The \HAEM system does not seem to learn well with naive alignment for this amount of data. The poorer performance of \HAEM whenever more training data are available is particularly obvious at the high-resource setting where the difference between \HACM and \HAEM is quite large.

At the low setting, both the HACM and HAEM
ensembles (Run 2 and Run 4) outperform the next
best competitor (LMU-02-0 with 46.59\%) by 0.23 and 1.94 percentage points
in average accuracy. The margin between Run 7 and the next best system is an impressive 4.02 percentage points.

At the medium setting, our best Run 7 also outperforms the next best competitor (LMU-02-0 with 82.64\%) with a small margin of 0.16 percentage points.  At the high setting, our best Run 7 loses against  UE-LMU-01-0 with a small margin of 0.20 percentage points.

The performance of our best system varies strongly across languages (Figure~\ref{fig:results-on-test-dev}). This is not only due to typological differences, but probably also because some languages have only inflection patterns for a single part-of-speech category (e.g. verbs in English) and other languages include nouns and adjectives (sometimes with very imbalanced class distributions).
Naive alignment generally works slightly better than smart alignment at the low setting (but sometimes fails detrimentally as in the case of Khaling, Navajo, or Sorani). For the medium and high settings, smart alignment strongly outperforms naive alignment for \HAEM, and a bit less so for \HACM. For a few languages such as Turkish, Haida or Norwegian-Nynorsk, naive alignment is consistently better than smart alignment.

As future work, we will experiment more with the \HAEM model and try to improve its capabilities for high-resource settings. One obvious option would be to use more fine-grained actions, for instance, directly learn substitutions for certain characters. This system would probably also profit from more consistent alignments. Even with smart alignments, we observe linguistically inconsistent character alignments that might also prevent useful generalizations.

\section{Related work}

Some task-specific work has been published after the 2016 edition of the SIGMORPHON Reinflection Shared Task \cite{CotterellKirov:2016} that dealt with 10 languages, providing training material roughly at the size of the high setting of the 2017 task edition (a mean training data set size of 12,751 samples with a standard deviation of 3,303). The winning system of 2016 \cite{Kann&Schutze2016a} showed that a standard sequence-to-sequence encoder-decoder architecture with soft attention \cite{BahdanauCB15}, familiar from neural machine translation, outperforms a number of other methods (as far as they were present in the task). 
Recently, \newcite{Aharoni&Goldberg2017} showed that hard monotonic attention works well when training data are scarce.
Their approach exploits the almost monotonic alignment between the lemma and its inflected form.
The \HACM model extends this work with a copying mechanism similar to the pointer-generator model of \newcite{Seeetal2017} and CopyNet of \newcite{GuCopyNet:2016}. In \HACM, the copying distribution, which is then mixed together with the generation distribution, is different: \newcite{Seeetal2017} employ the soft-attention distribution whereas \newcite{GuCopyNet:2016} use a separately learned distribution. Our \HACM model uses a simpler copying distribution that puts all the probability mass on the currently attended character. 
The logic of the \HAEM model is similar to that of SIGMORPHON 2016's baseline which uses a linear classifier over hand-crafted features to predict edit actions.
\newcite{Grefenstetteetal2015} extend an encoder-decoder model with neural data structures to
better handle natural language transduction. \newcite{Rastogietal2016} present a neural finite-state approach to string transduction.

\section{Conclusion}
In this large-scale evaluation of morphological inflection generation, we show that a novel neural transition-based approach can deal well with an extreme low-resource setup. For a medium size training set of 1K items, \HACM works slightly better. With abundant data (10K items), encoder/decoder architectures with soft attention are very strong, however, \HACM achieves a comparable development set performance.

For optimal results, the ensembling of different system runs is important. We experiment with different ensembling strategies for eliminating bad candidates.
%
At the low setting (100 samples), our best system combination achieves an average test set accuracy of 50.61\% (an average Levenshtein distance (LD) of 1.29), at the medium setting (1K samples) 82.8\% (LD 0.34), and at the high setting (10K samples) 95.12\% (LD 0.11). 


\section*{Acknowledgement}

We would like to thank the SIGMORPHON organizers for the exciting shared task and Tanja Samard\v{z}i\'{c} and two anonymous reviewers for their helpful comments. Peter Makarov has been supported by European Research Council Grant No. 338875.

\bibliographystyle{acl_natbib}
\bibliography{acl2017}

\end{document}